# Seeing the Forest Despite the Trees:
# Large Scale Spatial-Temporal Decision Making

**Mark Crowley**
Computer Science Department
University of British Columbia
cs.ubc.ca/∼crowley

**John Nelson**
Dept. of Forest Resources Management
University of British Columbia

**David Poole**
Computer Science Department
University of British Columbia
cs.ubc.ca/∼poole

## Abstract

We introduce a challenging real-world planning problem where actions must be taken at each location in a spatial area at each point in time. We use forestry planning as the motivating application. In Large Scale Spatial-Temporal (LSST) planning problems, the state and action spaces are defined as the cross-products of many local state and action spaces spread over a large spatial area such as a city or forest. These problems possess state uncertainty, have complex utility functions involving spatial constraints and we generally must rely on simulations rather than an explicit transition model. We define LSST problems as reinforcement learning problems and present a solution using policy gradients. We compare two different policy formulations: an explicit policy that identifies each location in space and the action to take there; and an abstract policy that defines the proportion of actions to take across all locations in space. We show that the abstract policy is more robust and achieves higher rewards with far fewer parameters than the elementary policy. This abstract policy is also a better fit to the properties that practitioners in LSST problem domains require for such methods to be widely useful.

## 1 INTRODUCTION

In some real world planning problems there are many actions to be taken in parallel over a spatial area. This is the case, for example, in urban planning when zoning different areas of a city for different uses. In infectious disease control, decisions need to be made about allocating medicine to thousands or millions of people spread across space based on need, cost, transportation or any number of other variables. In forestry planning, decisions come down to whether to cut each tree or not, or to perform some other activity at every point in the forest. We call problems of this form Large Scale Spatial-Temporal (LSST) planning problems.

After further motivating the problem with details from the example of forestry planning we introduce a general definition of LSST planning as a reinforcement learning (Sutton and Barto, 1998) problem and discuss the properties a solution needs to possess. We demonstrate how policy gradients (Williams, 1992) can satisfy many of these properties for LSST problems. We compare two policy formulations: an explicit policy that identifies each location in space with parameters for controlling the actions taken and an abstract policy that represents the proportion of actions that will be taken across the entire space. We show that this abstract policy produces better results with far fewer parameters and we argue that the level of abstraction it uses more closely matches the level needed by human planners in forestry or other LSST domains that would utilize this method to aid in planning.

## 2 FORESTRY BACKGROUND

Forestry planning as it is practiced in British Columbia will be used as our motivating example throughout this paper. Forestry is a very important industry in British Columbia generating 12% of the province's GDP and employing around 200,000 people. Large regions of forest of up to several hundred thousand hectares are licensed by the government to forestry companies to cut trees and sell lumber. The government places many constraints on management activities: setting a maximum annual allowable cut, specifying areas that are off limits, specifying spatial constraints to avoid a high density of cut areas and to protect wildlife migration routes and habitats. Violations of these constraints are enforced with large fines and possible revocation of licence.

Suppose you are the head forester in charge of planning for one of these companies. Your interest is to maximize your return and minimize your fines incurred. This can be achieved by providing a steady supply of logs over the long term and maintaining a healthy forest. The forest is divided up into many small spatial regions we will call *cells* and you must decide for every cell whether to clear-cut the whole cell, cut a portion of the trees or do nothing in that cell this year.

One major impact on forest health is insect infestation such as the Mountain Pine Beetle (MPB)(Eng et al., 2004). MPB are tiny beetles that burrow under the bark of pine trees laying eggs, cutting off nutrients and leaving a deadly blue fungus that kills the tree. MPB are an endemic species however, in recent decades, a lack of cold winters and the large number of older trees resulting from years of forest fire suppression have provided the MPB population with the conditions they need for an explosive epidemic. Cutting down trees before a brood spread can kill the beetles but the rice-sized beetles are hard to detect until a year or two after an attack. That is when the thousands of killed trees are easily spotted by their distinctive red color. The infestation is devastating the forests of British Columbia, wiping out over 50% of the harvestable pine in the past 15 years and shows no sign of stopping at the provincial or national borders.

## 2.1 SOLUTION CONSIDERATIONS

In a complex domain such as forestry there are many researchers who have developed sophisticated simulation models for different elements of forestry planning from tree growth to MPB growth. The explicit transition models underlying these simulations are too complex and varied to be used as conditional probabilities so we must rely on the simulations themselves as black boxes that provide a future forest state given some proposed set of actions and the current state. Generating simulations is expensive, so we need to treat all simulation data as a precious resource to be used as effectively as possible.

In many LSST planning domains a distinction is made between *strategic*, *tactical* and *operational* planning. Operational planning refers to the immediate implementation of a low level plan (eg. which particular trees to cut in which order). Tactical planning covers mid-sized regions over medium timescales of less than 20 years. Tactical plans take into account local conditions (eg. where to build roads to access the forest, assigning workers, scheduling cutting in different areas). Strategic planning takes place at the highest levels, focussing on properties of the entire landscape, spatial constraints and total rewards into the long term future over decades or centuries. Some strategic considerations in forestry are : maintaining the proportion of trees within an age class, balancing employment between regions, satisfying spatial constraints, reducing overall pest levels. The total number of cells in a landscape can range from 100-100,000. In this paper we focus on strategic planning, as this is the level where effective use of large amounts of data can have the greatest impact and it is an important area of research in Forestry planning.

The result of strategic planning is a *strategic policy* which is concerned primarily with the proportions of actions taken across the landscape and their impact on long term value. The strategic policy does not express which actions to take in particular cells in the landscape. This is due to the fact that over the level of the entire landscape there are many states that the reward function does not distinguish between. Consider a reward function based purely on the number of trees cut and constrained by a maximum allowable cut. There are many ways to achieve the maximum value that involve different assignments of actions to particular cells. These distinctions are not relevant as long as a policy can be defined to achieve the maximum value. In practice, the person designing the strategic policy often does not even have the authority to specify the lowest level action choices (eg. "Clear-cut cell 1582") as these choices are made by experts on the ground based on their local context *in accordance with* the strategic policy.

We thus have two major requirements for a good LSST planning solution:

1. the method can efficiently find a high value policy without having an explicit transition model for the simulation

2. we want a strategic policy that does not commit to more detail than necessary in order to maximize reward

In the following sections we give a general definition of LSST planning problems and show how to use policy gradients to achieve these requirements.

## 2.2 CURRENT SOLUTIONS IN FORESTRY

Many existing planning solutions in forestry have relied heavily on assuming spatial independence between cells in a forest. The deterministic optimization models often used, such as linear programming, break down when faced with uncertainty about the state of the forest and cannot use any information about spatial relations between cells. This is a problem in forestry (J.P.Kimmins et al., 2005) since MPB breaks the assumption of spatial independence and adds uncertainty to the problem. MPB can fly between nearby cells in the forest, so the immediate neighbourhood is always relevant when planning which trees to cut in order to reduce the spread of the pests and quickly salvage trees killed by them.

Other solution methods commonly used in forestry planning are simulation modelling and meta-heuristics. Simulation modelling is an interactive approach where the user specifies maps, constraints and preferences for various actions and the software carries out a simulation while choosing actions consistent with the constraints and preferences. Some example simulation tools are ATLAS (http://www.forestry.ubc.ca/atlas-simfor) and SELES (Fall et al., 2003). The results from these simulation planners are often then fed through other tools for analysis after which the user can alter the parameters of the simulation and run it once again. These simulations will be a useful black box to be used by higher level RL planning techniques.

The final set of methods in common use are stochastic local search methods such as tabu search, genetic algorithms and simulated annealing (Pukkala and Kurttila, 2005). The general approach is to predefine fixed plans that could be applied to a cell over the entire time horizon. The search proceeds to assign one of these plans to each cell, evaluate the outcome and make local improvement steps. Simulated annealing has offered the best hope for integrating spatial relations of these methods but uncertainty is generally not dealt with in a significant way.

Uncertainty is also introduced to the otherwise relatively predictable growth of trees by the fact that the exact location and severity of the MPB infestation is unknown until a year or two after an attack. Increasingly, there are efforts in forestry planning to improve modelling of uncertainty and complex, dynamic processes in the forest, such as fire or pest infestations (Baskent and Keles, 2005). This work contributes to those efforts by translating this specific domain into a general planning problem that can be approached with recent advances from the artificial intelligence community.

## 3    LSST DEFINITIONS

A landscape is partitioned spatially into a set of *cells* $C$. We assume these cells are disjoint and completely cover the area of the landscape. Cell partitioning remains constant over the planning horizon $T$.

Each cell, $c \in C$, at each timestep $t \in [0, T]$, has a state, $s \in S$. Each cell-state is a column vector of real numbers $s[f]$ for all cell-features $f \in F$. These features describe different aspects of the cell such as elevation, the number of trees in the cell, the number of trees per age class and the number of MPB present in the cell. It is also possible to model spatial features in each cell which take into account attributes of neighbouring cells. Such spatial features provide a simple way to include some relevant spatial information in local features of a cell. One such feature we will use in our experiments is an aggregated count of the number of MPB that will be invading the cell from all neighbouring cells.

The *landscape-state*, $\mathbf{s}$, represents the combined state[1] of all of the cells in the landscape as a function, $(\mathbf{s} : C \to S)$. The landscape-state at a particular timestep $t$ and cell $c$ is denoted $\mathbf{s}_t[c]$.

Each cell has an action, $a$, taken from the set of *cell-actions*

---

[1] Variables or functions refering to the entire landscape of cells will be set in bold.

$A$ which in this simplified forestry problem consists of: "clear-cut", "thin trees" and "do nothing". Similarly to states, the *landscape-action* is a function $(\mathbf{a} : C \to A)$ representing the combined actions in all cells in the landscape and referred to for a particular timestep $t$ and cell $c$ as $\mathbf{a}_t[c]$.

An LSST planning problem is defined as an *MDP* $\langle \mathbf{S}, \mathbf{A}, r, P \rangle$ where $\mathbf{S}$ and $\mathbf{A}$ are the sets of all possible landscape states and actions, $r$ is a reward function and $P$ is the state transition model. The transition model $P(\mathbf{s}_{t+1}|\mathbf{s}_t, a_t)$ will generally not be available in LSST problems in any explicit form. Instead, we assume there is access to simulations of the domain developed by domain experts. These external simulators return a new state when given the current landscape-state and landscape-action. By making a series of calls to the simulator we can construct a *trajectory*, $k$, of states and actions across all timesteps[2],
$k = \langle \mathbf{s}_0^k, \mathbf{a}_0^k, \mathbf{s}_1^k, \mathbf{a}_1^k, \ldots \rangle$.

The reward function, $r(\mathbf{s}_t, \mathbf{a}_t, \mathbf{s}_{t+1})$ returns a real number representing the reward received for the actions taken across the landscape $\mathbf{s}_t$. The reward may contain local cell components (eg. the value of cut trees), spatial components (eg. constraints on the number of contiguously cut cells) and even landscape-wide components (eg. a penalty for the total number of MPB present). Other important constraints in forestry are the annual allowable cut (AAC), as well as upper and lower bounds on the proportion of the forest in a different age classes (eg. no more than 20% of the forest is less than 10 years old). The total discounted reward of a trajectory is $R(k) = \sum_t \gamma^t r(\mathbf{s}_t^k, \mathbf{a}_t^k, \mathbf{s}_{t+1}^k)$ with a constant discount factor $\gamma \in [0, 1]$.

A *partially observable MDP* (POMDP) is defined as above except that the states are now hidden and there is an additional set, $O$, of observations about the states. The probability distribution $P(\mathbf{o}_{t+1}|\mathbf{a}_t, \mathbf{s}_{t+1})$ models how likely the observation is given the most recent action and the state that resulted from that action. Although LSST planning problems are partially observable in general, for this paper we focus on the fully observed problem.

### 3.1   POLICY DEFINITION

A cell-policy, $\pi$, is a distribution over actions for a given cell-state. The probability of taking action $a$ in a cell that is in state $s$ is given by $\pi(s, a, \theta)$. The policy parameters, $\theta$, are an $A \times F$ matrix of real numbers used to define the policy as a Gibbs distribution of weighted state features:

$$\pi(s, a, \theta) = \frac{e^{\theta[a]s}}{\sum_{b \in A} e^{\theta[b]s}} \qquad (1)$$

Where $\theta[a]$ is a vector of feature weights combined as a dot product with the cell-state feature vector, $s$.

---

[2] When it is not relevant, the trajectory $k$ will be dropped from state and actions names.

### 3.2 LANDSCAPE POLICY

A general landscape-policy, $\pi(\mathbf{s}, \mathbf{a}, \boldsymbol{\theta})$, can be defined in terms of $\pi$ as the joint probability of choosing all of the cell-actions in $\mathbf{a}$ given a set of cell-states, $\mathbf{s}$. The landscape-parameters $\boldsymbol{\theta}$ define parameters for each local cell-policy. The landscape-policy is computed as the product of the probabilities for all cells given by the appropriate cell-policies. We will provide two parameterizations for the landscape-policy, $\boldsymbol{\pi}^C$ and $\boldsymbol{\pi}^1$, shown below. The first formulation, $\boldsymbol{\pi}^C$, defines an explicit policy by maintaining separate parameters, $\boldsymbol{\theta}^C : C \to \theta$, for each cell and each time step.

$$\boldsymbol{\pi}^C(\mathbf{s}, \mathbf{a}, \boldsymbol{\theta}^C) = \prod_{c \in C} \pi(\mathbf{s}[c], \mathbf{a}[c], \boldsymbol{\theta}^C[c]) \quad (2)$$

The second formulation, shown in (3), describes an abstract policy where a single set of parameters, $\boldsymbol{\theta}^1$, is used for all cells in the landscape at that timestep.

$$\boldsymbol{\pi}^1(\mathbf{s}, \mathbf{a}, \boldsymbol{\theta}^1) = \prod_{c \in C} \pi(\mathbf{s}[c], \mathbf{a}[c], \boldsymbol{\theta}^1) \quad (3)$$

These two formulations will be used within a general policy gradient algorithm and compared.

## 4 POLICY GRADIENTS

Policy gradient (PG) methods seek to find optimal policies by following the gradient with respect to the policy parameters of a function describing the value of the current policy. PG researchers have recently achieved significant gains in the types of reinforcement learning problems that can be solved (Kersting and Driessens, 2008; Riedmiller et al., 2007; Sutton et al., 2000; Williams, 1992). PG methods require stochastic, parameterized policies and work well when the state space is very large and the transition dynamics are not available. These properties match well with LSST planning problems so PG methods seem a promising place to start looking for solutions.

Policy gradient algorithms are founded on the observation that the expected value of a stochastic policy can be computed using previously sampled trajectories by weighting the rewards actually received during each trajectory by the probability of that trajectory given the current policy (Sutton et al., 2000) (Riedmiller et al., 2007):

$$V^{\boldsymbol{\theta}} = E\left[p(k|\boldsymbol{\theta})R(k)\right] = \int_0^\infty p(k|\boldsymbol{\theta})R(k)dk \quad (4)$$

Where the probability of a trajectory $k$ is:

$$p(k|\boldsymbol{\theta}) = p(\mathbf{s}_0) \prod_t P(\mathbf{s}_t^k | \mathbf{s}_{t-1}^k, \mathbf{a}_{t-1}^k) \boldsymbol{\pi}(\mathbf{s}_t^k, \mathbf{a}_t^k, \boldsymbol{\theta}_t) \quad (5)$$

As stated earlier, in LSST planning problems we will generally be given a black box simulator rather than the transition model $P$. However, it turns out that computing the gradient of the value function with respect to the policy parameters, $\nabla_{\boldsymbol{\theta}} V^{\boldsymbol{\theta}}$, does not require knowing $V^{\boldsymbol{\theta}}$ or $P$ (Riedmiller et al., 2007; Sutton et al., 2000):

$$\begin{aligned} \nabla_{\boldsymbol{\theta}} V^{\boldsymbol{\theta}} &= \nabla_{\boldsymbol{\theta}} \int p(k|\boldsymbol{\theta})R(k)dk = \int \nabla_{\boldsymbol{\theta}} p(k|\boldsymbol{\theta})R(k)dk \\ &= \int p(k|\boldsymbol{\theta}) \nabla_{\boldsymbol{\theta}} \log p(k|\boldsymbol{\theta}) R(k) dk \\ &\approx \frac{1}{|K|} \sum_{k \in K} \nabla_{\boldsymbol{\theta}} \log p(k|\boldsymbol{\theta}) R(k) \end{aligned} \quad (6)$$

Where $K$ is the finite set of sampled trajectories. The log trajectory likelihood, $\nabla_{\boldsymbol{\theta}} \log p(k|\boldsymbol{\theta})$, needed in (6) can be computed using (5) to be:

$$\begin{aligned} \nabla_{\boldsymbol{\theta}} \log p(k|\boldsymbol{\theta}) &= \nabla_{\boldsymbol{\theta}} \log p(\mathbf{s}_0) + \nabla_{\boldsymbol{\theta}} \sum_t \log P(\mathbf{s}_t^k | \mathbf{s}_{t-1}^k, \mathbf{a}_{t-1}^k) \\ &\quad + \nabla_{\boldsymbol{\theta}} \sum_t \log \boldsymbol{\pi}(\mathbf{s}_t^k, \mathbf{a}_t^k, \boldsymbol{\theta}_t) \\ &= \sum_t \nabla_{\boldsymbol{\theta}} \log \boldsymbol{\pi}(\mathbf{s}_t^k, \mathbf{a}_t^k, \boldsymbol{\theta}_t) \end{aligned} \quad (7)$$

If we choose to use $\boldsymbol{\pi}^C$ as the landscape policy then the final gradient of the policy value becomes:

$$\begin{aligned} \nabla_{\boldsymbol{\theta}} V^{\boldsymbol{\theta}} &\approx \frac{1}{|K|} \sum_k \sum_t \nabla_{\boldsymbol{\theta}} \log \boldsymbol{\pi}^C(\mathbf{s}_t^k, \mathbf{a}_t^k, \boldsymbol{\theta}_t^C) R(k) \\ &= \frac{1}{|K|} \sum_k \sum_t \nabla_{\boldsymbol{\theta}} \log \prod_c \pi(\mathbf{s}_t^k[c], \mathbf{a}_t^k[c], \boldsymbol{\theta}_t^C[c]) R(k) \\ &= \frac{1}{|K|} \sum_k \sum_t \sum_c \nabla_{\boldsymbol{\theta}} \log \pi(\mathbf{s}_t^k[c], \mathbf{a}_t^k[c], \boldsymbol{\theta}_t^C[c]) R(k) \end{aligned} \quad (8)$$

The basic policy gradient algorithm then involves two main steps, *generating* samples and *updating* the policy. First, a new sample trajectory, $k$, is generated using the current policy parameters, $\boldsymbol{\theta}$, and this trajectory is used to compute $\nabla_{\boldsymbol{\theta}} V^{\boldsymbol{\theta}}$. Then, the policy parameters are updated by following the gradient of the policy value:

$$\boldsymbol{\theta}' = \boldsymbol{\theta} + \mu \nabla_{\boldsymbol{\theta}} V^{\boldsymbol{\theta}} \quad (9)$$

Where $\mu$ is a learning rate that controls the size of the policy update steps. This learning rate is notoriously difficult to choose as it needs to scale with the magnitude of the derivative. Riedmiller et al. (2007) describe a few techniques called *optimal base-lining* and *Rprop* to counteract these difficulties which we use. We describe these methods briefly here.

### 4.1 REDUCING THE VARIANCE OF $\nabla_{\boldsymbol{\theta}} V^{\boldsymbol{\theta}}$

The varying magnitude of $R(k)$ can lead to high variance in the estimate of $\nabla_{\boldsymbol{\theta}} V^{\boldsymbol{\theta}}$, which will impede learning. Part of the variance can be removed by subtracting a constant *baseline* $b$ from each occurrence of $R(k)$ in equations (4) and (8). This is valid since $\nabla_{\boldsymbol{\theta}} \int p(k|\boldsymbol{\theta})dk = \nabla_{\boldsymbol{\theta}} 1 = 0$ (Riedmiller et al., 2007). The optimal baseline for our problem (shown here for $\boldsymbol{\pi}^C$) is computed for each policy parameter $\theta[\alpha, f]$ for every $\alpha \in A$ and $f \in F$ as follows:

$$b_t[\alpha, f] =$$

$$\frac{\sum_k \left[\sum_t \sum_c \nabla_{\alpha f} \log \pi(\mathbf{s}_t^k[c], \mathbf{a}_t^k[c], \boldsymbol{\theta}_t^C[c])\right]^2 R(k)}{\sum_k \left[\sum_t \sum_c \nabla_{\alpha f} \log \pi(\mathbf{s}_t^k[c], \mathbf{a}_t^k[c], \boldsymbol{\theta}_t^C[c])\right]^2} \quad (10)$$

Another technique used to improve policy gradient performance is called Rprop (Riedmiller et al., 2007) which replaces scaled updating using the full gradient as in eq (9) with an update-value, $\Delta_{\boldsymbol{\theta}}$, which has the *same direction* as $\nabla_{\boldsymbol{\theta}} V^{\boldsymbol{\theta}}$ but a magnitude that is unrelated to the gradient. The magnitude of $\Delta_{\boldsymbol{\theta}}$ is similar to the magnitude of the parameters in $\boldsymbol{\theta}$ and is updated incrementally based on the progress of the policy search. To update the policy we compute $\boldsymbol{\theta}' = \boldsymbol{\theta} + \Delta_{\boldsymbol{\theta}}$ and then increment the value of $\Delta_{\boldsymbol{\theta}}$ in the appropriate direction based on the gradient. See (Riedmiller et al., 2007) for more details.

### 4.2 SOLVING LSST PROBLEMS USING POLICY GRADIENTS

The formulation for $\nabla_{\boldsymbol{\theta}} V^{\boldsymbol{\theta}}$ in (8) requires us to know $\nabla \log \pi(s, a, \theta)$. Our choice of policy parameterization allows us to express this analytically. We compute the partial derivative $\nabla_{\alpha f} V^{\boldsymbol{\theta}}$ with respect to parameter $\theta[\alpha, f]$ for every $\alpha \in A$ and $f \in F$:

$$\nabla_{\alpha f} \log \pi(s, a, \theta) = \nabla_{\alpha f} \log \left(\frac{e^{\theta[a]s}}{\sum_{b \in A} e^{\theta[b]s}}\right)$$

$$= \nabla_{\alpha f} \log e^{\theta[a]s} - \nabla_{\alpha f} \log \sum_b e^{\theta[b]s}$$

$$= \nabla_{\alpha f} \theta[a]s - \frac{\nabla_{\alpha f} \sum_b e^{\theta[b]s}}{\sum_b e^{\theta[b]s}}$$

$$= \nabla_{\alpha f} \theta[a]s - \frac{\sum_b e^{\theta[b]s} \nabla_{\alpha f} \theta[b]s}{\sum_b e^{\theta[b]s}}$$

This partial derivative will be different depending on whether the action for this cell, $a$, matches the action associated with parameter being differentiated, $\alpha$. Since all the policy parameters are independent we know that for any action $a \in A$ and feature $g \in F$:

$$\nabla_{\alpha f} \theta[a, g] = \begin{cases} 1 & \text{if } \alpha = a \text{ and } f = g \\ 0 & \text{otherwise} \end{cases} \quad (11)$$

This allows us to simplify $\nabla_{\alpha f} \log \pi(s, a, \theta)$ to:

$$s[f](1 - \pi(s, \alpha, \theta)) \quad : \text{ if } \alpha = a \quad (12)$$
$$-s[f]\pi(s, \alpha, \theta) \quad : \text{ if } \alpha \neq a \quad (13)$$

### 4.3 LSST POLICY GRADIENT ALGORITHM

Combining all of these elements together we arrive at the policy gradient algorithm that iteratively generates new trajectories and updates the policy based on the current set of trajectories:

**Algorithm:** LSST-PG($s_0$)
initialize $\boldsymbol{\theta}$ randomly
$\Delta_{\boldsymbol{\theta}} = 0.1; K = \emptyset$
**repeat** *maxSamples* **times**
　// Sample new trajectory
　$\langle \mathbf{s}, \mathbf{a}, R \rangle$ = generateTrajectory($\mathbf{s}_0, \boldsymbol{\theta}$)
　$K = K \cup \langle \mathbf{s}, \mathbf{a}, R \rangle$
　// Update policy
　**update** $b$ as in (10)
　$\nabla_{\boldsymbol{\theta}} V^{\boldsymbol{\theta}} = \frac{1}{|K|} \sum_k \sum_t (R(k) - b_t) \nabla_{\boldsymbol{\theta}} \log \boldsymbol{\pi}(\mathbf{s}_t^k, \mathbf{a}_t^k, \boldsymbol{\theta}_t)$
　**update** $\Delta_{\boldsymbol{\theta}}$ using $\nabla_{\boldsymbol{\theta}} V^{\boldsymbol{\theta}}$ as in sec 4.1
　$\boldsymbol{\theta} = \boldsymbol{\theta} + \Delta_{\boldsymbol{\theta}}$
**return** $\boldsymbol{\theta}$

generateTrajectory($\mathbf{s}_0, \boldsymbol{\theta}$)
$R = 0$
**for** $t = 0$ **to** $T$ **do**
　$\langle \mathbf{a}_t, r_t, \mathbf{s}_{t+1} \rangle$ = runSim($\mathbf{s}_t, \boldsymbol{\theta}_t$)
　$R = R + \gamma^t r_t$
**return** $\langle \mathbf{s}, \mathbf{a}, R \rangle$

runSim($\mathbf{s}_t, \boldsymbol{\theta}_t$)
**foreach** $c$ *in* $C$ **do**
　// sample action distribution
　$\mathbf{a}_t[c] \sim \pi(\mathbf{s}_t[c], \boldsymbol{\theta}_t^C[c])$　　(or $\pi(\mathbf{s}_t[c], \boldsymbol{\theta}_t^1)$)
$\mathbf{s}_{t+1}$ = externalSimulator($\mathbf{s}_t, \mathbf{a}_t$)
$r_t$ = reward($\mathbf{s}_t, \mathbf{a}_t, \mathbf{s}_{t+1}$)
**return** $\langle \mathbf{a}_t, r_t, \mathbf{s}_{t+1} \rangle$

We set the initial value of the gradient update-value, $\Delta_{\boldsymbol{\theta}}$ to a value of 0.1 which has been found to be reasonable for many problems (Riedmiller et al., 2007). The main loop repeats until *maxSamples* is reached which is simply an upper bound on the number of trajectories to sample. Some other condition could easily be used such as a measure of the current convergence of the gradient. Note also that the sample and update steps are independent and could be run varying numbers of times or in parallel.

## 5　TWO ALGORITHM VARIANTS

In section 2.1, we outlined two major requirements of a good LSST planning solution, dealing with the lack of an explicit transition model and defining a strategic policy that does not overcommit to too much low level detail.

Policy gradients provide a way to satisfy the first requirement as they do not require a model and only follow the gradient of the policy value. One fairly obvious approach is to define a landscape policy using $\pi^C$ which maintains parameters for every aspect of the state space with $\theta_t^C[c]$ defined for each and every cell. We will call this algorithm LSST-PG$^C$ and it is simply the LSST-PG algorithm shown above where $\pi^C$ fulfills the function of $\pi$ in the code.

### 5.1 ABSTRACT ACTIONS

The algorithm LSST-PG$^C$ has two major problems. First, it gives us an enormous number of parameters to search over with $|A|\times|F|\times T\times|C|$ dimensions. As we mentioned earlier, $|C|$ could be on the order of 100,000 whereas $|A|$, $|F|$ and $T$ are generally less than 100. This enormous space makes convergence to an optimal policy very difficult.

The second problem is that LSST-PG$^C$ does not give us a strategic policy but instead a very low level operational policy. The optimal strategic policy should not distinguish between particular cells. The policy should treat cells interchangeably and define a pattern of actions that gives the proportion of cells each action will be applied to across the landscape. We do not want to require a commitment to particular actions for particular cells in our strategic policy.

The policy $\pi^C$ is, in essence, too focussed on the 'trees' to ever see the 'forest' and find these patterns of actions. To achieve a strategic policy we instead propose to use a single, stochastic action for the entire landscape and a single set of parameters $\theta_t^1$ for each step in time. This is the policy $\pi^1$ shown in equation (3). We define a second algorithm, LSST-PG$^1$, where the role of $\pi$ in LSST-PG is fulfilled by $\pi^1$ instead of $\pi^C$. We also need to alter the sampling line in the third method `runSim` to $a_t[c] \sim \pi(s_t[c], \theta_t^1)$.

This shift to one set of parameters seems minor, but its impact is profound. By optimizing $\pi^1$ we will be learning how to act on abstract cells that could occur anywhere in the landscape. While the policy is still defined for each cell, it now does not distinguish between cells based on their identity. All cells are treated equally based on their state features. Recall that cell features can also take into account information from neighbouring cells such as MPB spread.

One way to think about the difference between $\pi^C$ and $\pi^1$ is by an analogy to time. A policy can be stationary or nonstationary with respect to time. A stationary policy defines one set of parameters for all timesteps. Similarly, $\pi^1$ is stationary with respect to space. This *spatially stationary policy* defines a distribution over actions based on cell features that apply to any cell in the landscape. Note that this spatially stationary policy is well-defined for any number of cells and thus has arbitrary scale much as a stationary policy has arbitrary scale in terms of planning horizons. A spatially stationary policy would have many advantages during planning, allowing us to easily change scale or apply a learned policy to different subregions of the landscape without modification.

## 6 EXPERIMENTS

The goal of our research is to develop a planning algorithm that can utilize existing simulations from LSST domains in a scalable way to find high value strategic policies. There are a great variety of simulators in forestry that each require extensive expertise to set up and integrate with. We decided for this stage of our research to develop our own simple forest simulator to evaluate the performance of gradient descent on this problem. Our simulator includes state features for the distribution of tree species and age classes, the level of MPB in a cell and its neighbouring cells. The dynamics include tree birth, growth and death, replanting of young trees after clearcutting, killing of trees by MPB and the spread of MPB to nearby cells year to year.

We implemented the two algorithms, LSST-PG$^1$ and LSST-PG$^C$, in Matlab and ran all tests on a dual processor Pentium 4 3.2GHz PC with 2GB RAM running Windows XP.

The initial landscape states were varied randomly around representative values for state features based on common distributions present in data for BC forests for tree species, tree age, MPB presence and other features.

The reward function assigns value to individual trees cut and penalizes various properties of the landscape state, such as, a quadratic penalty on the deviation from a desired tree density for the entire landscape, linear penalties for overcutting and for the number of trees killed by MPB, and base costs for maintaining the forest (salaries, license fees, etc.) to inhibit a strategy of no cutting at all. Note that the reward function is not equally well defined at all points. The "Do Nothing" policy $\langle DoNothing = 1.0, ClearCut = 0.0, Thin = 0.0\rangle$ is a bad policy to follow in our model, as it is all cost and no revenue, but it is actually much worse than the reward indicates. Modellers are not willing to even assign a utility to situations where the entire industry ceases to exist or, similarly, where all of the trees are cut down and the ecosystem is totally destroyed. Rewards can be defined accurately within "reasonable" regions of policy space, but they must still be defined at all points to serve as a signal to be used during policy search.

## 7 RESULTS

Figure 1 shows a typical result for the total reward received by the two algorithms. The reward is shown for each trajectory sample and is averaged over 20 trials for a small problem with 5 cells and 5 timesteps. Each trial sampled 200 trajectories and updated the policy after every 5 samples using all trajectories sampled up to that point. The initial policy for each trial was specified by uniform weights across all state features combined with an initial action distribution for the action components of the parameters: $\langle DoNothing = 1.0, ClearCut = 0.0, Thin = 0.0\rangle$. Ini-tially all timesteps (and cells) will have the same action distribution before they begin diverging.

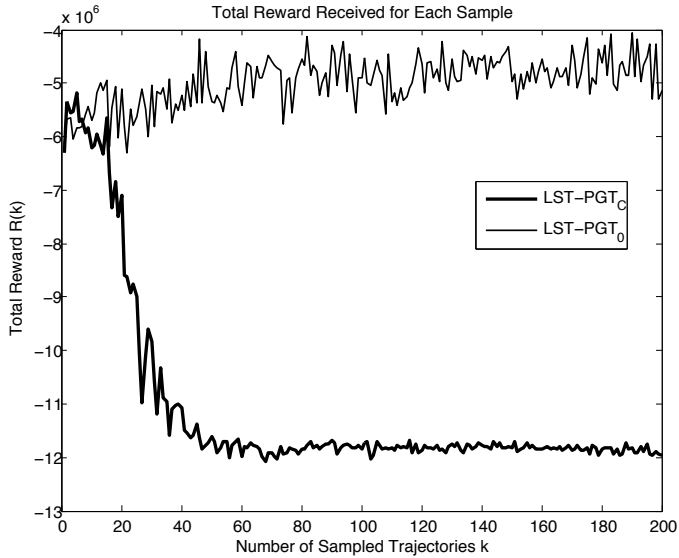

Figure 1: Total reward received average over 20 trials for LSST-PG$^C$ and LSST-PG$^1$ on 5 cells with 5 timesteps after 200 samples with policy updates every five samples. Initial policy was $\langle DoNothing = 1.0, ClearCut = 0.0, Thin = 0.0\rangle$.

LSST-PG$^1$ consistently finds higher value policies than LSST-PG$^C$. The abstract policy of LSST-PG$^1$ is more robust across multiple trials whereas LSST-PG$^C$ fixes onto a deterministic set of action assignments to particular cells that is tuned to the start state for each trial.

Figure 2 shows the initial and final policies for the two algorithms on a single trial of a 20 cell planning problem. The initial policy in this trial was set to $\langle DoNothing = .8, ClearCut = .15, Thin = .05\rangle$ for both algorithms. The actions that are actually taken are not exactly the same even with identical initial parameters because of the different policy structures and stochastic choices being made but the final policies are very different. LSST-PG$^1$ has found a policy that does even less cutting than the initial policy but that cuts more in timesteps seven and eight to achieve a higher value.

The LSST-PG$^C$ algorithm has found a policy with a much higher proportion of cutting. This policy is deterministic, each cell at each timestep always has the same action taken over many different trajectory samples. LSST-PG$^C$ cannot break out of this policy, even though it is incurring major penalties for overcutting, because the value of a policy is based on weighting rewards by the liklihood of past trajectories under the current policy. This makes a deterministic policy that doesn't change between samples very attractive. Once a deterministic policy is found, diverging on some cell will only lower the expected value of the policy. Algorithm LSST-PG$^1$ does not have this problem since there are fewer deterministic policies in which to get stuck and they all have very low reward (all "Do Nothing, all "Cut" etc.).

Figure 3 shows the gradients of the combined parameters for each action in a policy for the same trial as in figure 2. After the LSST-PG$^C$ policy converges to a narrow range of rewards the gradient begins converging. For LSST-PG$^1$, the variation in the gradient drops significantly once a 'good' policy is found.

The LSST-PG$^1$ algorithm runs about three times faster than LSST-PG$^C$ on the same number of trials and trajectories. This is not surprising since both algorithms sample actions for every cell and timestep while LSST-PG$^C$ uses more memory and time managing the large number of parameters. The actual runtimes for LSST-PG$^1$ range from 2 minutes for a 5 cell, 5 timestep problem, up to 230 minutes for a 10 timestep, 30 cell problem. With the current implementation we estimate that solving a problem with a couple thousand cells would take about two days. Our implementation has a lot of room for efficiency improvements but other advances will be needed to improve speeds even further. For realistic problem sizes of hundreds of thousands of cells, forestry planners currently expect runtimes of tens of minutes (for linear problems) or up to several hours (for 'meta-heuristic' solutions).

## 8  RELATED AND FUTURE WORK

We have used our own forest simulator here to keep implementation simple. To improve realism it would be best to switch to a simulator in use by forestry planners. The tools used for simulation planning such as those discussed in section 2.2 could be used for this. We are working with researchers in forestry to integrate our algorithm with more of their own data and simulations to experiment at the larger scales needed for results to be useful for Forestry planning experts.

Our work builds upon research on model-free reinforcement learning (RL)(Sutton and Barto, 1998) and policy gradient methods (Riedmiller et al., 2007). Sutton et al. (2000) described the basis for using policy gradients within an RL framework. The algorithm presented here includes some extensions to basic PG such as reward baselining and Rprop. More advanced PG techniques are available such as natural gradients, which have been shown by Riedmiller et al. (2007) to significantly improve performance by computing the reward baseline using the Fisher information matrix of the gradient. Another obvious next step is to fully model the uncertainty about the current state and use policy gradients to solve a POMDP version of the LSST problem. Also, the independence of the update policy and generate trajectory steps would make parallelization of the algorithm straightforward.

Outside of policy gradient methods there is Least Square

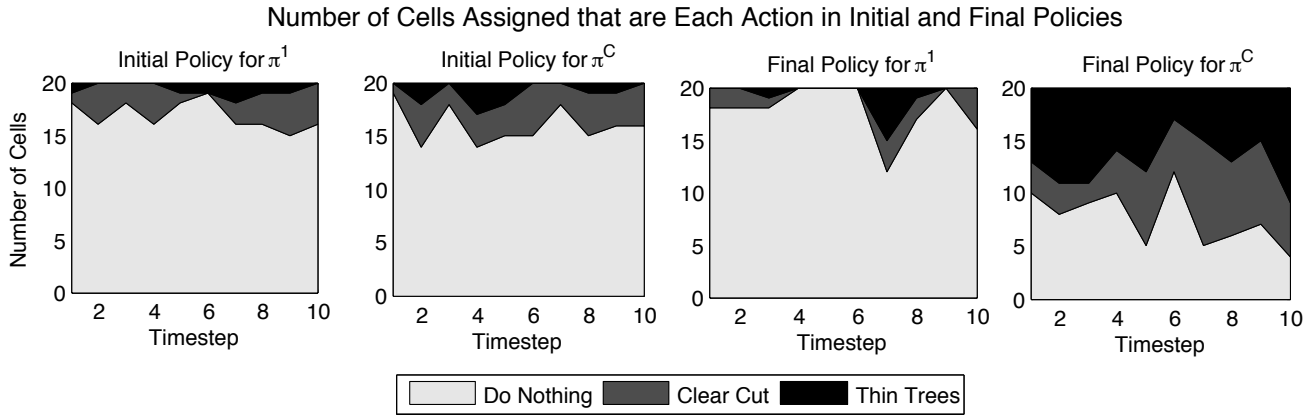

Figure 2: In each cell the available actions are $\langle \text{``DoNothing''}, \text{``ClearCut''}, \text{``ThinTrees''} \rangle$. The number of cells assigned each of these actions are shown as areas. The initial and final policies after optimization for both LSST-PG$^1$ and LSST-PG$^C$ are shown. This trial used 200 sampled trajectories using, 10 timesteps and 20 cells. The initial policy was set to $\langle DoNothing = .8, ClearCut = .15, Thin = .05 \rangle$ for both algorithms.

Policy Iteration (LSPI) (Lagoudakis and Parr, 2001), another RL approach that uses a parameterized policy and learns without a transition model by using a stored history of sampled trajectories. As the policy changes the value of the policy can be recomputed based on these trajectories.

LSST problems also have many similarities to multi-agent planning problems as each cell can be viewed as an agent while we are seeking to optimize a joint reward based on the actions of all agents. Guestrin et al. (2002) applied LSPI to multi-agent problems to find an initial estimate of the value function and specify a policy in proportion to the values of each action. New research into decentralized (PO)MDPs (Seuken and Zilberstein, 2008) brings together various threads in coordinated multi-agent planning into one language. LSST planning problems could be a rich problem domain for this new field. However, many multi-agent methods assume that the value function is a linear combination of the local value functions of individual agents. In LSST problems this assumption does not always hold as there are nonlocal constraints on actions across the landscape such as landscape cut quotas.

The explicit policy, $\pi^C$, is overly detailed and unwieldy for real world planning. The abstract policy, $\pi^1$, is at a more appropriate, strategic level of abstraction but it is merely the other extreme end of a spectrum of policies. In between are varying levels of abstraction that could be defined by using more than one set of policy parameters applied to groups of cells. These groups could be learned from clustering of features or by hierarchical decomposition of cell-state space by iteratively adding new features to define groups of similar cells. Another idea along these lines to explore is using multiple weighted policies where the weights determine which policies are applied to which cells and these weights are part of the learning process.

Spatial relations between cells are a major component of LSST problems which we have addressed here with simple aggregation features from neighbouring cells. This method could be expanded with more complex relational aggregators or by adding new variables modelling relations between groups of cells and represented in the reward and policy functions. Recently, Kersting and Driessens (2008) introduced a non-parametric policy gradient approach that might be useful for LSST planning. Their method uses a gradient tree boosting approach for learning policies in relational domains.

The long timescales used in LSST problems ensure that long term plans will not be followed blindly for any length of time. Thus, planning over time periods of varying lengths, such as is done in the SMDP literature (Barto and Mahadevan, 2003), could be useful. There may be enormous gains available to be made by dynamically allowing greater abstraction of policies, models and time granularity as time progresses.

## 9 CONCLUSIONS

In this paper we have introduced Large Scale Spatial-Temporal planning problems, which are very challenging instances of general planning where states and actions are spatially divided into components. We have described how to apply RL techniques to these problems and demonstrated one way to use policy gradient methods to find good policies in a simulated forestry planning problem. We showed that use of a spatially stationary policy formulation greatly reduces the parameter space to be searched and improves the value of the resulting policy.

We hope that raising awareness about this particular set of problems will benefit both the UAI research community and the many researchers and planners in real-world planning domains with LSST structures who are looking for a way to make their very complex problems more manageable.

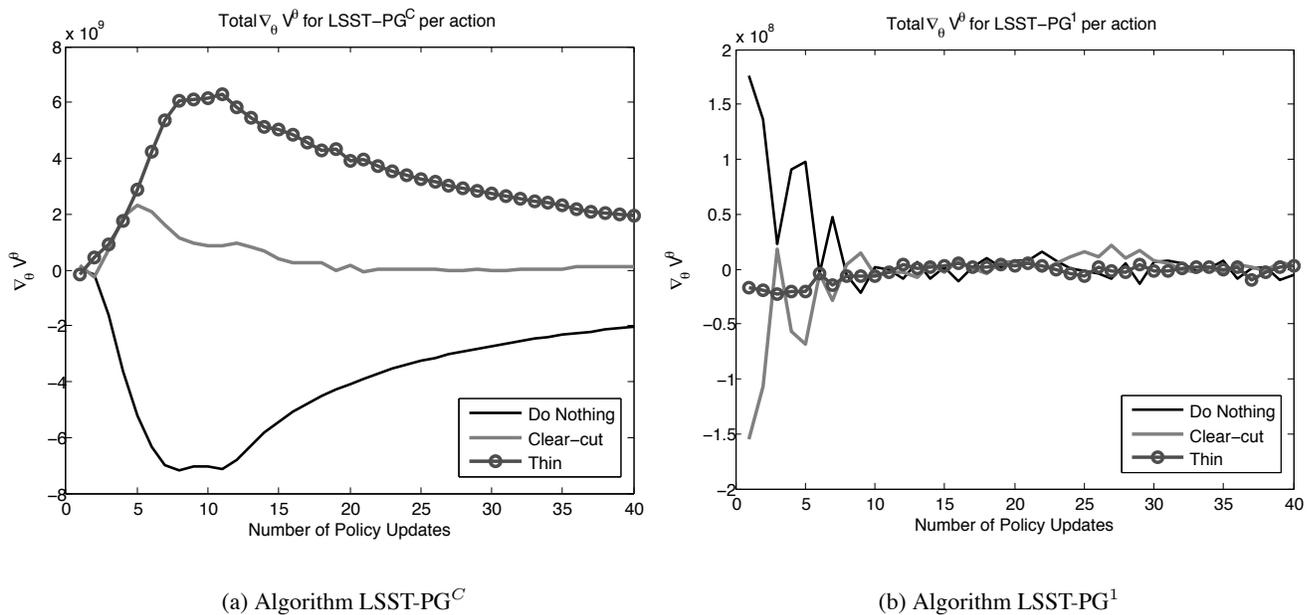

(a) Algorithm LSST-PG$^C$          (b) Algorithm LSST-PG$^1$

Figure 3: Policy gradients for parameters relating to each action, summed across all cells and timesteps for one trial with 200 sampled trajectories, 20 cells and 10 timesteps.

**Acknowledgements**

We would like to thank Matt Hoffman for his help with PG methods and feedback received from Peter Carbonetto, Michael Chiang, Albert Jiang, Jacek Kisyński and the very helpful advice from the anonymous reviewers.